\documentclass[10pt,twocolumn,letterpaper]{article}

\usepackage[pagenumbers]{cvpr} % To force page numbers, e.g. for an arXiv version

\usepackage{booktabs}
\usepackage{multirow}

\definecolor{cvprblue}{rgb}{0.21,0.49,0.74}
\usepackage[pagebackref,breaklinks,colorlinks,allcolors=cvprblue]{hyperref}

\def\confName{CVPR}
\def\confYear{2026}

\title{Pushing the Limits of Distillation-Based Continual Learning via Classifier-Proximal Lightweight Plugins}

\author{Zhiming Xu$^{1,2}$, Baile Xu$^{1,2}$, Jian Zhao$^{3}$, Furao Shen$^{1,2,\dag}$, Suorong Yang$^{1,3}$\\
$^{1}$ National Key Laboratory for Novel Software Technology, Nanjing University, China\\
$^{2}$ School of Artificial Intelligence, Nanjing University, China \\
$^{3}$ Department of Computer Science and Technology, Nanjing University, China\\
$^{4}$ School of Electronic Science and Engineering, Nanjing University, China\\
{\tt\small \{york\_z\_xu,sryang\}@smail.nju.edu.cn}, \tt\small\{blxu,jianzhao,frshen\}@nju.edu.cn
}

\begin{document}
\maketitle
\begin{abstract}
Continual learning requires models to learn continuously while preserving prior knowledge under evolving data streams. Distillation-based methods are appealing for retaining past knowledge in a shared single-model framework with low storage overhead. However, they remain constrained by the stability-plasticity dilemma: knowledge acquisition and preservation are still optimized through coupled objectives, and existing enhancement methods do not alter this underlying bottleneck. To address this issue, we propose a plugin extension paradigm termed Distillation-aware Lightweight Components (DLC) for distillation-based CL. DLC deploys lightweight residual plugins into the base feature extractor's classifier-proximal layer, enabling semantic-level residual correction for better classification accuracy while minimizing disruption to the overall feature extraction process. During inference, plugin-enhanced representations are aggregated to produce classification predictions. To mitigate interference from non-target plugins, we further introduce a lightweight weighting unit that learns to assign importance scores to different plugin-enhanced representations. DLC could deliver a significant 8\% accuracy gain on large-scale benchmarks while introducing only a 4\% increase in backbone parameters, highlighting its exceptional efficiency. Moreover, DLC is compatible with other plug-and-play CL enhancements and delivers additional gains when combined with them.  
\end{abstract}


\maketitle

\section{Introduction}
%In recent years, deep learning has achieved remarkable success across a variety of domains~\cite{jaimes2007multimodal,ye2019learning,chen2021large,chen2022learning,yang2024entaugment}. 
In recent years, deep learning has achieved remarkable success across a broad spectrum of multimedia applications~\cite{yang2024entaugment,de2024towards,gui2024navigating,wu2025gui,ye2025safedriverag}. 
However, most existing models are built on the assumption that data distributions and semantic categories remain fixed after training. 
In real-world multimedia scenarios, data streams evolve continuously, and novel categories emerge over time~\cite{gomes2017survey}, making static training paradigms increasingly inadequate. 
Meanwhile, naively adapting models to newly arrived data often requires additional specific modules, leading to inefficient model growth and limited scalability. 
Continual learning (CL)~\cite{zhou2024class,liangloss,yang2025supervised} therefore plays a crucial role in enabling deep learning models to continuously adapt to non-stationary data while preserving prior knowledge.

Existing CL methods~\cite{zhou2024class} can be broadly categorized into rehearsal-based, restriction-based, and expansion-based approaches. 
Rehearsal-based methods maintain a small replay buffer by storing a limited number of representative exemplars from earlier learning stages~\cite{lopez2017gradient,rolnick2019experience,de2021rep}. Other categories of methods likewise adopt rehearsal mechanisms to retain exemplars from previous learning stages, thereby further improving model accuracy. 
Expansion-based CL methods~\cite{mallya2018piggyback,serra2018overcoming,xu2018reinforced} allocate stage-specific architectural blocks, creating isolated parameter spaces for newly acquired knowledge while freezing previously learned parameters. This strategy alleviates catastrophic forgetting and preserves plasticity during subsequent learning stages. However, rather than updating a single model progressively, these methods expand the network and store, for each learning stage, an additional full~\cite{yan2021dynamically} or partial (about 70\%)~\cite{zhou2022model} feature extractor, incurring substantial storage costs. For instance, with ResNet-18, each new learning stage adds about 11.7MB of parameters, roughly equivalent to storing 3,998 images, which is about twice the replay buffer capacity under standard CL settings. This overhead can limit their practicality for continual learning.

\begin{figure}[!t]
  \centering
  \includegraphics[width=3in,keepaspectratio]{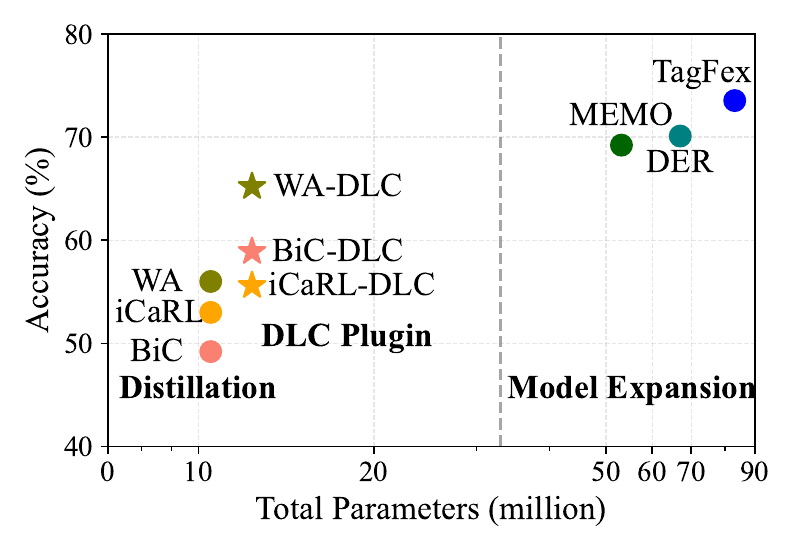}
  \caption{Parameter-accuracy comparison of different CL methods on ImageNet-100. DLC achieves the best balance between parameter efficiency and accuracy compared to prior distillation-based and SOTA expansion-based methods. iCaRL-DLC, WA-DLC, and BiC-DLC mean integrating DLC into the corresponding methods, respectively.}
  % All methods utilize the ResNet-18 as initialization. DLC serves as a plugin extension, which attains a favorable accuracy gain with minimal parameter overhead, increasing parameters by only 4.39\% to the original network.}
    \label{fig:params}
\end{figure}

Restriction-based methods \cite{nguyen2017variational, ahn2019uncertainty} aim to mitigate catastrophic forgetting by constraining model updates during new learning stages.
A representative example is knowledge distillation~\cite{rebuffi2017icarl, simon2021learning}, which preserves previous knowledge by aligning the current model's output with its earlier versions. 
These methods perform CL on a single model, leading to substantially lower storage overhead. As a result, these approaches are practically attractive in real-world settings. However, they inevitably introduce a trade-off \cite{de2021continual}: the model must balance adherence to past knowledge with adaptation to new data, since the distillation and knowledge acquisition are coupled objectives whose gradients can conflict.
This stability-plasticity dilemma often becomes the key bottleneck for distillation-based approaches. Many enhancement methods \cite{wu2019large,zhao2020maintaining,gao2025maintaining} have been proposed to improve their accuracy, but most of them mainly introduce additional training objectives, leaving the core stability-plasticity dilemma fundamentally unresolved, leaving performance governed by the same trade-off. A natural way to reduce this interference is to expand extra parameters. However, scaling up the feature extractor is quickly parameter-prohibitive, and expanding multiple extractors breaks the shared single-model predictor that distillation relies on.
This exposes a new quandary to CL: \textit{How can we break the stability–plasticity dilemma by introducing a small set of auxiliary parameters, while remaining friendly to distillation-based methods?}

To address this challenge, we propose the Distillation-aware Lightweight Components (DLC), a plug-and-play extension paradigm for CL. We revisit parameter-efficient tuning modules in transfer learning and study them under the training dynamics of distillation-based CL, where the backbone is not a knowledge-rich pretrained model but a dynamically evolved learner trained over continual stages.
We attach several lightweight plugins to the classifier-proximal layer of the base feature extractor, i.e., its final feature layer before classification. These plugins perform specific semantic calibration by injecting residual corrections into classifier-proximal features. When adding each plugin, we first follow the underlying baseline to train the backbone while keeping all plugins frozen, and then freeze the backbone to train only the plugin introduced for the current stage. This decoupled update strategy trains each plugin independently, enabling DLC to be deployed into the baseline without interference.
During inference, the model activates all plugins to extract their feature representations, which are aggregated for final classification. However, plugins that are not aligned with the input sample may contribute noisy or weakly informative residuals. We introduce a lightweight weighting unit before the classifier, which automatically assigns importance weights to different plugin outputs.
Experimental results on multiple benchmarks show that DLC consistently improves replay and distillation-based baselines. For example, DLC achieves up to 8\% accuracy gains on ImageNet-100 with ResNet-18 and on CIFAR-100 with ResNet-32, while increasing the total parameter count by only 4\% to 30\%. Under the same memory budget, DLC also outperforms existing state-of-the-art CL baselines.

Our contributions are summarized as follows:
\begin{itemize}
\item We propose DLC, a plug-and-play plugin extension framework that deploys lightweight residual plugins at classifier-proximal layers in continually updated backbones, enabling effective semantic correction while preserving compatibility with distillation-preserved knowledge.

\item We introduce a lightweight weighting unit that adaptively suppresses irrelevant plugin activations based on learned importance, leading to more stable and accurate continual learning.

\item Extensive experiments show that DLC significantly boosts the performance of various CL methods and remains compatible with other plug-and-play enhancement techniques. Meanwhile, under a fixed memory budget, the DLC-enhanced distillation method can surpass state-of-the-art expansion-based CL methods.

\end{itemize}

\section{Related Work}
\subsection{Continual learning}
Continual learning primarily aims to prevent the model from forgetting acquired knowledge. Existing methods tackle this challenge from multiple perspectives. Rehearsal-based methods~\cite{li2025re,aghasanli2025prototype} replays representative exemplars from previous classes. Regularization \cite{bian2024make} or distillation-based methods~\cite{rebuffi2017icarl,wu2019large,zhao2020maintaining,he2024gradient} limit overly aggressive updates on new data. Expansion-based methods train and retain task-specific full~\cite{yan2021dynamically} or partial~\cite{zhou2022model} feature extractors, and further integrate contrastive objectives~\cite{zheng2025task} or pre-trained models~\cite{zhou2024expandable,sun2025mos} for better representations. These methods often exhibit a performance–memory trade-off: approaches with higher overall performance typically require storing more replay samples or allocating and maintaining additional model parameters through expansion.

\subsection{Distillation-Based CL}
Distillation-based CL trains the model on new data while regularizing it to match the predictions of the previous version. A wide range of distillation-based approaches have been explored, including combining distillation with replay~\cite{rebuffi2017icarl}, incorporating bias correction~\cite{wu2019large} or weight alignment~\cite{zhao2020maintaining}, momentum-based distillation~\cite{michel2024rethinking}, and plug-and-play enhancement methods~\cite{bian2024make,gao2025maintaining,li2025c}. Distillation remains appealing as it introduces minimal overhead, can leverage unlabeled incoming samples, and fits a simple single-model training loop, making it broadly applicable in practical settings~\cite{wang2024improving,nori2025federated, zeng2025terrasap}.
However, the feature extractor’s fixed capacity imposes a fundamental constraint: the model must balance stability and plasticity by trading off new-data learning and distillation losses, ultimately capping achievable performance.

\begin{figure*}[!ht]
  \centering
  \includegraphics[width=5.95in,keepaspectratio]{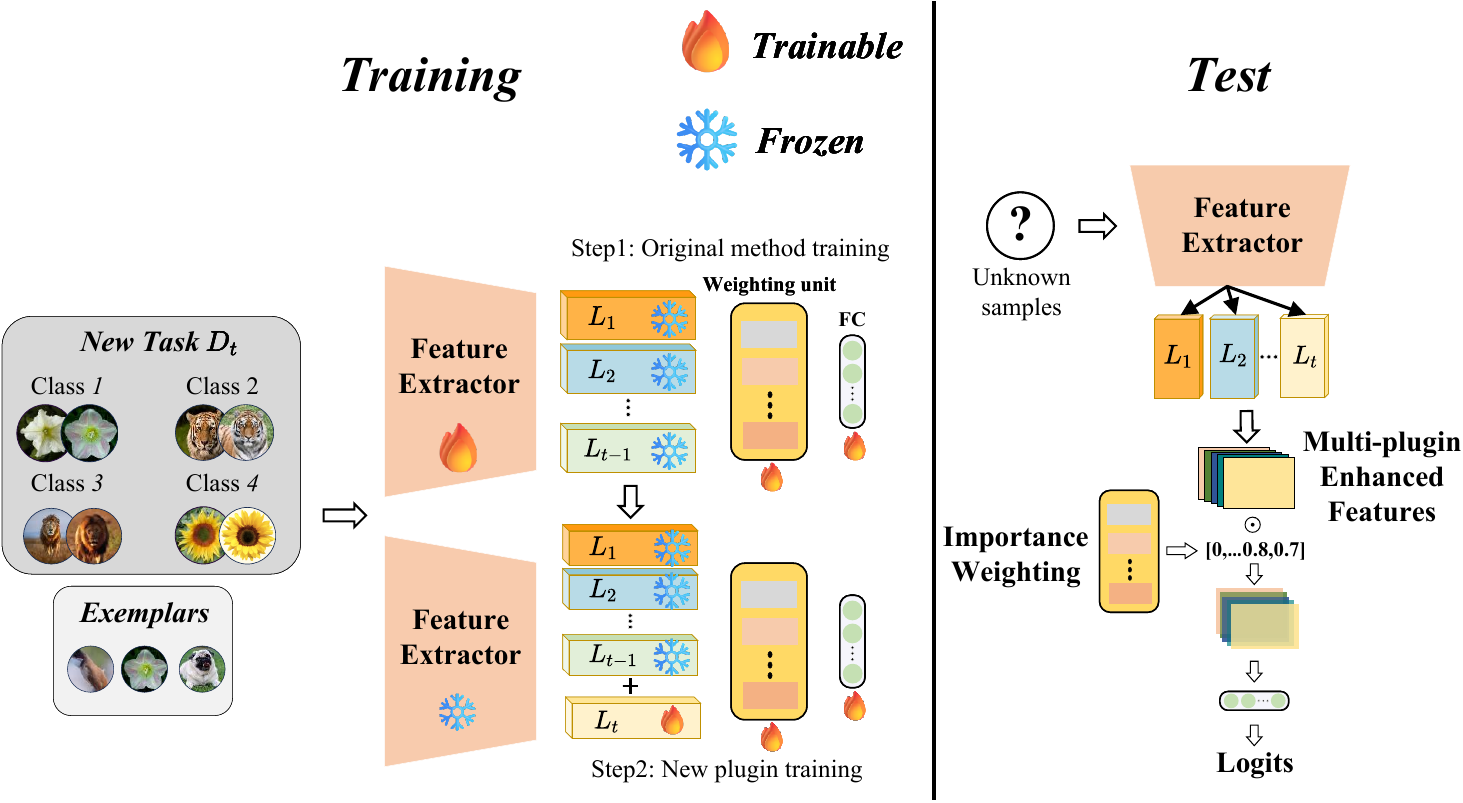}
  \caption{The proposed DLC framework. \textbf{Left:} Training. When a new $D_t$ arrives, a dedicated plugin $L_t$ is created. First follows the base method to train the backbone, and then freezes the backbone to train $L_t$. \textbf{Right:} Test.    Using all plugins to enhance the feature extractor representations, then reweight by a weighting unit, concatenated, and classified.}
    \label{fig:dlcwork}
\end{figure*}

\section{Preliminaries}
\subsection{Problem Formulation}
A CL model aims to learn new data while retaining previous knowledge from the continuously evolving data streams $D_1, D_2,\cdots, D_t$, where $D_i = \{(\mathbf{x}_j,y_j) \}_{j=1}^{n_i}$ containing $n_i$ instances. At each learning stage $t$, the current data stream $D_t$ is treated as a task, and training is performed using only $D_t$ with a few exemplars from the previous stage \cite{rebuffi2017icarl}. After the model advances to the next learning stage, all earlier stages, except for the stored exemplars, will become inaccessible. When there are test samples for inference, the model needs to make predictions based on the label spaces of all the tasks it has encountered. We do not provide the task ID of the sample to the model, making this a more challenging class-incremental continual learning \cite{de2021continual} setup.

For any stage $t$, the model has learned $D = D_1 \cup D_2 \cup \cdots \cup D_t$ tasks, 
We aim to find a model $f(\mathbf{x}): \mathcal{X} \rightarrow \mathcal{Y}$ that minimizes
empirical risk across all learned data, where $\mathcal{X}$ and $\mathcal{Y} = Y_1 \cup Y_2 \cdots \cup Y_t$ represent the sample's feature and label spaces, respectively. The effectiveness of the model is evaluated by the test dataset  $D^{test} = D_1^{test} \cup D_2^{test} \cup \cdots \cup D_t^{test}$, which can be expressed in Eq.\eqref{eq:s3e1}:
\begin{equation}
    \label{eq:s3e1}
    f^{*}(\mathbf{x}) =\arg\min _{f \in \mathbb{H}}\mathop{\mathbb{E}}\nolimits_{(\mathbf{x},y) \in D^{test}}[\mathbb{I}(f(\mathbf{x})\neq y)] ,
\end{equation}
where $\mathbb{H}$ is the hypothesis space of model and $\mathbb{I} (\cdot)$ denotes the indicator function. 

\subsection{‌Knowledge in Distillation-Based CL}
\label{sec32}
Considering the classic CL scenario with only a single feature extractor. The objective function used in non-CL settings is the cross-entropy loss $\mathcal{L}_{\text{CE}}$, while distillation-based CL methods incorporate an additional restriction term $\mathcal{L}_{\text{KD}}$ that enforces the preservation of knowledge acquired from previous data. This composite objective function can be formally expressed as:
\begin{equation}
    \label{eq:s3e2}
    \mathcal{L} = \mathcal{L}_{\text{CE}} + \mathcal{L}_{\text{KD}}.
\end{equation}
As a canonical example of distillation-based methods, we examine logit-based distillation, which ensures consistency between the logit distributions produced by the old and new models for known classes. Specifically, let $\hat{q}(\boldsymbol{x})$ and $q(\boldsymbol{x})$ represent the logits of the teacher and student models. The cross-entropy version of $\mathcal{L}_{\text{KD}}$ is defined as:
\begin{equation}
\mathcal{L}_{\text{KD}} = -\sum_{i} \hat{q}_i(\boldsymbol{x}) \log (q_i(\boldsymbol{x})),
\end{equation}
and the KL divergence version is defined as:
\begin{equation}
\mathcal{L}_{\text{KD}} = \tau^{2} \sum_{i} \hat{q}_i(\boldsymbol{x}) \log (\frac{\hat{q}_i(\boldsymbol{x})}{q_i(\boldsymbol{x})}),
\end{equation}
where the $\tau$ is the temperature scalar. Most of these methods are complemented by a replay buffer that stores a subset of representative exemplars from previous data. These exemplars are interleaved with data from the current stage during training to help maintain performance on past data. Specifically, in the conventional setting, the model is trained only on the newly arrived data $D_t$. With the replay buffer, the training data is supplemented with $D_{\text{exp}}$ retrieved from the buffer, forming a combined training set $D_{\text{train}}= D_t \cup D_{\text{exp}}$.
Although the model's parameters are dynamically updated throughout the continual learning process, the use of knowledge distillation and the replay buffer ensures that a portion of knowledge from previous data remains preserved. 
Replay ensures the model continues to minimize the empirical risk on a representative subset of past data, anchoring the parameters in regions consistent with previous data. Concurrently, knowledge distillation functions as a functional regularizer, ensuring that the mappings for old classes remain stable by aligning the output logits (or distributions) with those of the prior model. 
Consequently, a model trained with replay and distillation preserves substantial knowledge of past classes, resulting in a single unified model that is broadly effective but suboptimal. We posit that the knowledge stored in the model can be efficiently activated with minor auxiliary parameters.

\section{The Proposed Method}
DLC proposes a plugin-level residual injection paradigm, as illustrated in Fig. \ref{fig:dlcwork}. We treat a distillation-based CL network as the base model that accumulates knowledge across each CL stage but remains suboptimal in accuracy, and then augment it with auxiliary plugins that inject residual knowledge into the extracted features to improve performance. 
The proposed DLC is detailed in the following subsections.
 
%%%%%%%%%%%%%
\subsection{Classifier-Proximal Plugin Deployment}
\label{sec41}
We introduce a plugin extension design for continual learning by maintaining an independent residual plugin for each learning stage. Specifically, for stage $t$, we attach a dedicated plugin $L_t$ for the feature extractor $\phi(\boldsymbol{x})$, and denote the resulting plugin-enhanced representation by $\phi^{L_t}(\boldsymbol{x})$. Rather than modifying the backbone weights directly, each plugin acts as a lightweight residual branch that performs stage-specific semantic correction on the extracted features. In this way, DLC complements the continually updated backbone with additional discriminative knowledge in a plug-and-play manner, while preserving the backbone as the main carrier of shared knowledge across stages.

DLC adopts a classifier-proximal deployment strategy, where plugins are attached to the final layer of the $\phi(\boldsymbol{x})$, i.e., the layer closest to the classifier. This placement is beneficial for two reasons. First, classifier-proximal features are already semantically mature and directly support classification, so residual correction at this stage can more effectively improve inter-class discriminability. Second, the injected perturbation is localized near the classifier and thus minimally disrupts the overall feature extraction process. Compared with shallow-layer intervention, such a design is more compatible with distillation-based objectives, since shallow-layer feature changes would otherwise propagate through the network and amplify output deviations, making output alignment between the old and current models harder to optimize.

The plugins and $\phi(\boldsymbol{x})$ are trained in a decoupled two-phase manner. When a new task $D_t$ arrives, we first follow the underlying baseline to train the backbone and classifier while keeping all plugins frozen. After this phase, the backbone is frozen, and only the newly introduced plugin $L_t$ is optimized on the same data. Once trained, $L_t$ is permanently frozen to preserve the acquired stage-specific knowledge. The plugin objective combines the cross-entropy loss with an auxiliary loss commonly adopted in expansion-style methods \cite{yan2021dynamically}:
\begin{equation}
    \label{eq:lora}
    \mathcal{L}_{\text{plg}} = \mathcal{L}_{\text{CE}} + \mathcal{L}_{\text{aux}}.
\end{equation}
This decoupled procedure allows plugins to learn stage-specific corrections without being directly entangled with distillation optimization, while also preventing plugin training from interfering with the baseline update of the backbone.

During inference, all learned plugins are activated to produce their corresponding plugin-enhanced representations. Because each plugin encodes a different stage-specific residual refinement on top of the shared backbone output, these representations provide diverse semantic cues with potentially complementary information. Their aggregation can thus be viewed as a classifier-proximal form of multi-stage feature reuse, in spirit similar to DenseNet \cite{huang2017densely} feature reuse, enriching the final prediction with knowledge accumulated across learning stages. For a linear classifier $\boldsymbol{W}$, the classification process can be written as:
\begin{equation}
    \hat{y} = \boldsymbol{W} [\phi^{L_1}(\boldsymbol{x}), \phi^{L_2}(\boldsymbol{x}), \ldots, \phi^{L_T}(\boldsymbol{x})],
\end{equation}
where $T$ denotes the total number of learned stages so far. Notably, the full feature extractor does not need to be executed $T$ times: since each plugin injects a residual after the last layer, the backbone activations before the deployed layer can be shared and computed once, avoiding redundant computation. 

For continual learning to remain practical in real-world scenarios, the goal is not to allocate an extractor-level or layer-level plugin for every learning stage, but to introduce a lightweight auxiliary module plugin with minimal storage overhead. Such a design naturally calls for parameter-efficient fine-tuning (PEFT) modules \cite{han2024parameter}. However, these modules have previously been studied mostly in transfer learning, where they are typically deployed on top of a knowledge-rich pretrained backbone. Distillation-based continual learning presents a fundamentally different regime: the backbone is not a static pretrained model with abundant prior knowledge, but a continually updated learner whose knowledge is progressively accumulated and preserved through replay and distillation. As discussed in Sec.~\ref{sec32}, knowledge distillation together with a replay buffer helps retain previously acquired knowledge, thereby constraining distribution drift during continual updates. This makes it possible to revisit PEFT modules in a dynamic backbone. Concretely, after training on stage $t$, the output deviation at layer $\ell$ is bounded by $K_\ell \Gamma_t$, where $K_\ell$ depends on hyperparameters such as the network architecture and distillation temperature, and $\Gamma_t$ reflects the distribution discrepancy induced by the adopted replay and distillation strategy. This bound directly limits the worst-case feature drift on the old-stage distribution, enabling stable optimization of PEFT modules in this dynamic regime. A detailed proof is provided in the supplementary material.

\subsection{LoRA-Style Plugins}
To instantiate each stage-specific plugin with minimal parameters, we adopt a LoRA~\cite{hu2022lora}-style low-rank residual parameterization. This design is well-suited for plugin deployment: the residual structure supports our decoupled training and inference aggregation since plugins can be independently trained, frozen, and combined without modifying the backbone. We emphasize that the backbone weights are kept intact throughout, and the low-rank component is introduced as an auxiliary module that adds residual to the feature. That is, the LoRA plugin does not replace or factorize the original weights, but injects additional knowledge via an additive low-rank branch, which aligns with our plug-and-play deployment setting.

Following the classifier-proximal deployment strategy in Sec.~\ref {sec41}, this plugin is attached to the final feature layer of $\phi(\boldsymbol{x})$; for CNN backbones, it is instantiated on the representation produced by the last convolutional layer. However, LoRA was originally proposed for Transformer models, while distillation-based CL methods are often built upon CNN-based backbones trained from scratch. We therefore provide a convolutional LoRA-style plugin instantiation for this setting, implemented as a lightweight low-rank residual branch with two convolutions.
For a convolutional layer with weight tensor $W_{\text{conv}} \in \mathbb{R}^{C_{\text{out}} \times C_{\text{in}} \times K \times K}$ and input $\boldsymbol{x} \in \mathbb{R}^{N \times C_{\text{in}} \times H \times W}$, the original forward pass is $\boldsymbol{h} = W_{\text{conv}} \ast \boldsymbol{x}$, where $\ast$ denotes the convolution operation. The LoRA plugin first processes the input with $\mathbf{A}$:
\begin{equation}
\boldsymbol{z} = \mathbf{A} \ast \boldsymbol{x},
\label{eq:conv_lora_step1}
\end{equation}
where $\mathbf{A} \in \mathbb{R}^{r \times C_{\text{in}} \times K \times K}$ projects the feature to a low-rank space. This is followed by $\mathbf{B}$:
\begin{equation}
\boldsymbol{h}_{\text{lora}} = \mathbf{B} \ast \boldsymbol{z} ,
\label{eq:conv_lora_step2}
\end{equation}
where $\mathbf{B}$ $\in \mathbb{R}^{C{\text{out}} \times r \times 1 \times 1}$ restores the channel dimensionality, and $\boldsymbol{h}_{\text{lora}}$ is added as a residual to the layer output. 
In summary, the forward pass of a convolutional layer equipped with the LoRA-style residual plugin is:
\begin{equation}
\boldsymbol{h}' = W_{\text{conv}} \ast \boldsymbol{x} + \frac{\alpha}{r}(\mathbf{B} \ast (\mathbf{A} \ast \boldsymbol{x})),
\label{eq:conv_lora_complete}
\end{equation}
where $\alpha$ is a scaling hyperparameter, and $r$ is the rank. Following \cite{hayou2024lora+}, $r$ and $\alpha$ are set to the recommended values of 8 and 16.

Compared to the model or layer-level expansion, instantiating the LoRA-style PEFT plugins can bring significant parameter advantages. Specifically, for a convolutional layer ($C_{in}$, $C_{out}$, $K \times K$), the space complexity of its parameters is $\Theta(C_{in} C_{out} K^2)$, while the plugin only occupies $\Theta(r (C_{in} K^2 + C_{out}))$. For expanding one layer of CNN over $T$ stages, considering that in a CNN in CL, $C_{out}$ is generally greater than or equal to $C_{in}$, we set $C_{in} = C_{out} = C$, and $K$ can be regarded as a constant (usually 3 or 7). Moreover, since the low rank $r$ is set to 8, it can be treated as a constant. Then, using LoRA-style plugin instantiation reduces the parameter-space complexity of expanding a single CNN layer from $\Theta(T C^2)$ to $\Theta(T C)$. Considering that $C$ is generally between 64 and 512, reducing its order of magnitude from quadratic to linear can substantially reduce the number of parameters.
%%%%%%%%%%%%%%%%%%%%%%%%%%%%%%%%%%%%%
\subsection{Dynamic Weighting Unit}
For any sample from stage $t$, all the plugins trained on other tasks $D_{i\neq t}$ are considered non-target plugins. 
Given that the replay buffer contains only a minimal fraction of the original data, these non-target plugins are not trained on the full $D_t$. In particular, plugins trained on stages 1 to $t-1$ never encounter samples from $D_t$. Consequently, the residuals extracted by these plugins could be largely irrelevant to the stage $t$, thus introducing uninformative features or even detrimental noise to the classifier. To mitigate their influence, we will create a weighting unit corresponding to concentrated representations when training on a new stage. 
This unit comprises two linear layers followed by a sigmoid function $\sigma(\cdot)$, denoted as $\omega(\mathbf{x}) = \sigma \Big( \mathbf{W}_2 \text{ReLU} \big( \mathbf{W}_1 \mathbf{x} \big) \Big), \mathbf{x} \in \mathbb{R}^{k}, \mathbf{W}_1 \in \mathbb{R}^{k \times d}, \mathbf{W}_2 \in \mathbb{R}^{d \times k}$, where $\sigma(\cdot)$ normalizes the output to the [0, 1] range.
During inference, we first compute the importance weights for the aggregated plugin-enhanced representation:
\begin{equation}
\boldsymbol{\omega}=
\omega\!\left(
[\phi^{L_1}(\boldsymbol{x}),\phi^{L_2}(\boldsymbol{x}),\ldots,\phi^{L_T}(\boldsymbol{x})]
\right),
\end{equation}
then perform inference using the reweighted representation:
\begin{equation}
\hat{y}=
\boldsymbol{W}
\left(
\boldsymbol{\omega}
\odot
[\phi^{L_1}(\boldsymbol{x}),\phi^{L_2}(\boldsymbol{x}),\ldots,\phi^{L_T}(\boldsymbol{x})]
\right).
\end{equation}
Where the $\odot$ denotes element-wise multiplication. The weighting unit can be regarded as a gate for the classifier. Jointly trained with the classifier, it helps the model adapt to the high-dimensional feature space and automatically assigns lower weights to uninformative or noisy features.

The automatically trained weighting unit fails to explicitly model the varying importance of representations from different stages.
For a sample $i$ in stage $t$, we denote the representations adapted by plugins trained on stages 1 to $t-1$ ($[\phi^{L_1}(\boldsymbol{x}),\phi^{L_2}(\boldsymbol{x}),\ldots,\phi^{L_{t-1}}(\boldsymbol{x})]$) as $\mathbf{h}^{(i)}_{\text{pre}}$ and those adapted by the remaining plugins as $\mathbf{h}^{(i)}_{\text{pos}}$, respectively.
Then $\mathbf{h}^{(i)}_{\text{pre}}$ should be assigned lower weights as corresponding plugins were not optimized with samples in $D_t$. 
Since the stage of each sample is unavailable during the CL testing, we cannot explicitly control the weighting unit based on stage identity. 
Instead, we employ a supervised regression training to guide the weighting unit to approximate the desired weighting scheme. We decompose the weight $\boldsymbol{\omega}^{(i)}$ for sample $i$ into $[\boldsymbol{\omega}^{(i)}_{\text{pre}},\boldsymbol{\omega}^{(i)}_{\text{pos}}]$
Then the decision process of the linear classifier can be rewritten as follows:
\begin{equation}
    \boldsymbol{q}^i = \mathbf{W}_{\text{pre}}(\boldsymbol{\omega}^{(i)}_{\text{pre}} \odot \mathbf{h}^{(i)}_{\text{pre}}) + \mathbf{W}_{\text{pos}}(\boldsymbol{\omega}^{(i)}_{\text{pos}} \odot \mathbf{h}^{(i)}_{\text{pos}}) .
\end{equation}
Formally, the contribution of $\mathbf{h}^{(i)}_{\text{pre}}$ to the classifier's logits can be expressed as $\mathbf{\xi}= \mathbf{W}_{\text{pre}} (\boldsymbol{\omega}^{(i)}_{\text{pre}} \odot \mathbf{h}^{(i)}_{\text{pre}})$. To generate effective weights, an importance-aware optimization objective could be incorporated into the training process, denoted as $\mathcal{L}_{\text{IA}}$. This loss function employs the Mean Squared Error (MSE), expressed as follows:
\begin{equation}
   \mathcal{L}_{\text{IA}} = \frac{1}{k} {\Vert \boldsymbol{\omega}-\boldsymbol{\omega}_{\text{ideal}} \Vert}_2,
\end{equation}
where the $\boldsymbol{\omega}_{\text{ideal}}$ is the ideal weight output, it has $\boldsymbol{\omega}^{(i)}_{\text{ideal-pos}} = 1$ and $\boldsymbol{\omega}^{(i)}_{\text{ideal-pre}} = 0$. 
In the ideal training case, the model has $\boldsymbol{\omega}^{(i)}_{\text{pos}} \to 1$ and $\boldsymbol{\omega}^{(i)}_{\text{pre}} \to 0$, then it has $\mathbb{E}[\mathbf{\xi}]\to0$ and $\text{Var}[\mathbf{\xi}]\to0$, so the $\mathbf{h}^{(i)}_{\text{pre}}$ have negligible influence on prediction logits.
Even if the ideal scenario is hard to achieve, the classifier can still produce more accurate predictions as long as the representations adapted by reliable plugins are assigned higher weights, while those adapted by non-target plugins are appropriately suppressed.
After enabling the weighting unit, $\mathcal{L}_{\text{IA}}$ is incorporated into the plugin training, and $\mathcal{L}_{\text{plg}}$ in Eq. \eqref{eq:lora} is extended as follows:
\begin{equation}
    \mathcal{L}_{\text{plg}} = \mathcal{L}_{\text{CE}} + \mathcal{L}_{\text{aux}} + \mathcal{L}_{\text{IA}}.
\end{equation}

\begin{table*}[!ht]
\begin{center}
\caption{Last performance $A_T$ and average performance $\overline{A}$ comparison.}
\scalebox{1}{
\begin{tabular}{l||cc|cc|cc||cc||cc}
\hline
\multirow{3}{*}{\textbf{Methods}} & 
\multicolumn{6}{c||}{\textbf{CIFAR-100}} & 
\multicolumn{2}{c||}{\textbf{Tiny-ImageNet}} & 
\multicolumn{2}{c}{\textbf{ImageNet-100}} \\
\cline{2-11}
& 
\multicolumn{2}{c|}{\textbf{B5 Inc5}} & 
\multicolumn{2}{c|}{\textbf{B10 Inc10}} & 
\multicolumn{2}{c||}{\textbf{B50 Inc10}} & 
\multicolumn{2}{c||}{\textbf{B40 Inc40}} & 
\multicolumn{2}{c}{\textbf{B50 Inc10}} \\
\cline{2-11}
& $\overline{A}$ & $A_T$ & $\overline{A}$ & $A_T$ 
& $\overline{A}$ & $A_T$ & $\overline{A}$ & $A_T$ 
& $\overline{A}$ & $A_T$  \\
\hline

Replay     & 58.28 & 38.20 & 56.82 & 39.31 & 51.08 & 40.88 & 51.46 &37.79 & 55.22 &43.16   \\ 
Replay-DLC & 58.76 & 38.69 & 59.24 & 41.81 & 53.72 & 43.61 & 51.65 & 38.66 & 59.76 & 49.62   \\ 
iCaRL      & 59.40 & 39.94 & 59.67 & 41.29 & 55.94 & 44.44 & 52.32 & 36.22 & 62.72 & 53.55 \\
iCaRL-DLC  & 59.54 & 41.63 & 60.06 & 44.31 & 56.61 & 46.35 & 53.28 & 39.20 & 65.03 & 54.06 \\
BiC        & 56.45 & 32.77 & 61.14 & 42.58 & 53.45 & 36.04 & 54.12 & 39.68 & 66.02 & 49.82 \\ 
BiC-DLC    & 58.27 & 41.07 & 61.87 & 43.65 & 54.46 & 46.23 & 54.86 & 43.09 & 71.53 & 57.24 \\
WA         & 62.72 & 45.96 & 65.82 & 51.80 & 65.04 & 56.94 & 53.68 & 39.69 & 65.53 & 56.55\\
WA-DLC     & 62.84 & 46.22 & 66.79 & 52.70 & 65.43 & 57.71 & 55.31 & 43.33 & 73.84 & 64.40 \\ 
MKD        & 56.86 & 36.87 & 55.73 & 36.83 & 46.68 & 36.17 & 49.34 & 34.64 & 65.64 & 54.68 \\
MKD-DLC    & 57.51 & 37.33 & 58.10 & 40.40 & 52.18 & 40.99 & 50.29 & 35.75 & 66.42 & 56.23   \\ 
DAKD       & 60.01 & 40.15 & 60.69 & 42.32 & 54.33 & 43.09 & 51.98 & 36.73 & 60.76 & 51.50 \\
DAKD-DLC   & 60.47 & 41.54 & 61.01 & 43.81 & 56.99 & 46.00 & 52.78 & 37.38 & 64.73 & 53.94   \\ 

\hline
Extra/Feat \#P (MB)  & 0.15 &0.46 & 0.07 &0.46 & 0.04 & 0.46  & 0.45 & 11.17  & 0.52 & 11.17\\ 
\hline
\end{tabular}
}
\label{tab:bc}
\end{center}
\end{table*}

\begin{figure*}[!t]
  \centering
  \subfloat[B10 Inc10 with AA]{
      \includegraphics[width=2.1in,keepaspectratio]{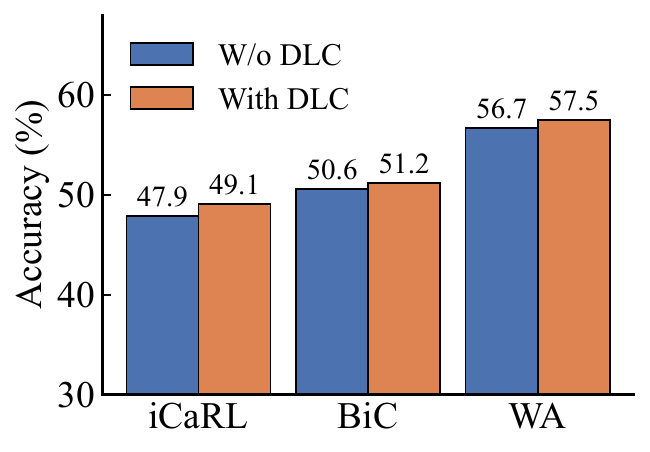}}
  \subfloat[B10 Inc10 with $\mathcal{L}_{\text{inter-intra}}$]{
      \includegraphics[width=2.1in,keepaspectratio]{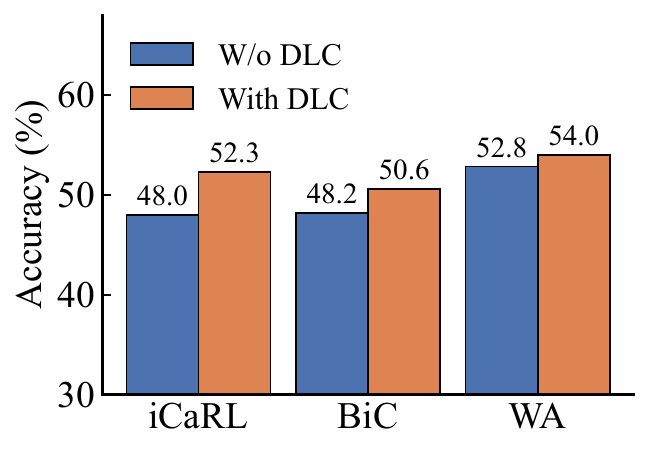}}
  \subfloat[B10 Inc10 with C-flat]{
      \includegraphics[width=2.1in,keepaspectratio]{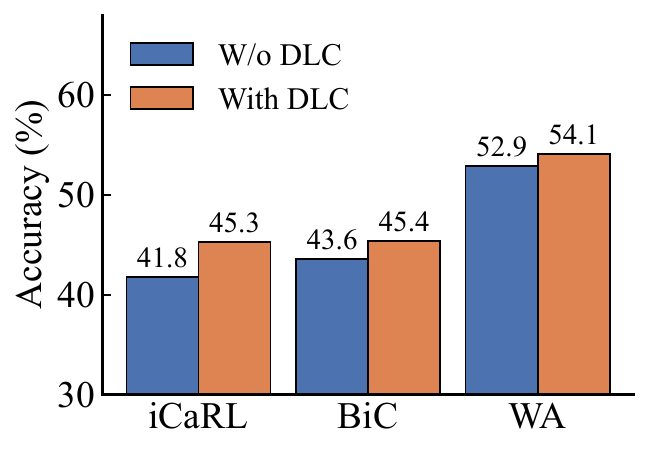}}
      
   \subfloat[B50 Inc10 with AA]{
      \includegraphics[width=2.1in,keepaspectratio]{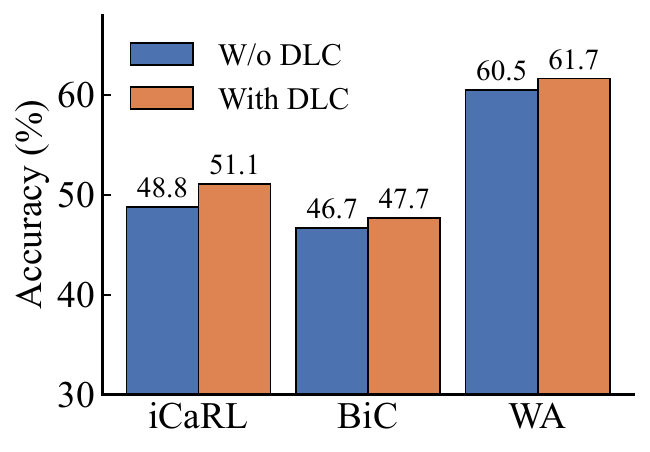}} 
   \subfloat[B50 Inc10 with $\mathcal{L}_{\text{inter-intra}}$]{
      \includegraphics[width=2.1in,keepaspectratio]{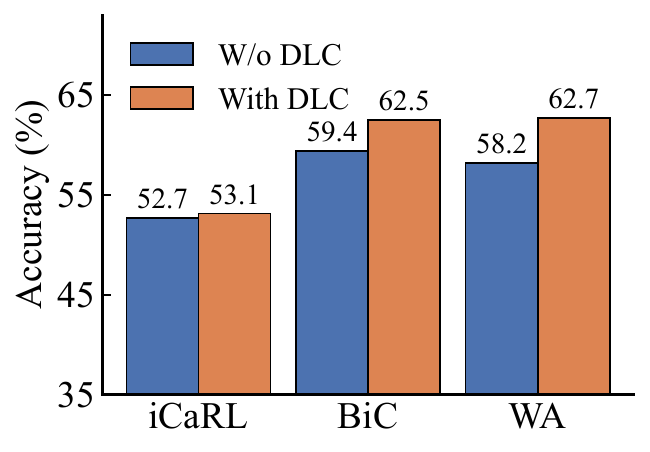}}   
   \subfloat[B50 Inc10 with C-flat]{
      \includegraphics[width=2.1in,keepaspectratio]{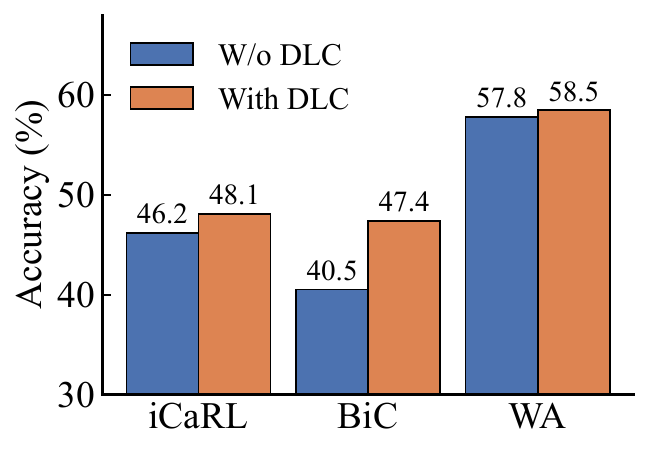}}   
  \caption{Comparison of $A_T$ with/without DLC when enhanced with AutoAugment (AA) , $\mathcal{L}_{\text{inter-intra}}$ and C-flat.}
  \label{fig:cmp1}
\end{figure*}

\section{Experiments}
This section conducts extensive experiments. Sec. \ref{sec:52} evaluates the validity of DLC on distillation-based baselines. Sec.~\ref{sec:cap} further evaluates DLC under challenging CL settings, including long-sequence, few-shot, and class-imbalanced scenarios. Sec. \ref{sec:mema} investigates the performance of SOTA CL methods under the memory-aligned case. Sec. \ref{sec:55} analyzes the additional inference overhead introduced by DLC in terms of FLOPs and latency.
%-------------------------------------------------------------------------
\subsection{Experimental Setup}
\textbf{Datasets and Split.} We chose three benchmark datasets commonly used in CL, including CIFAR-100 \cite{krizhevsky2009learning}, Tiny-ImageNet-200 \cite{vinyals2016matching}, and the large-scale ImageNet-100 \cite{deng2009imagenet}. Following the ‘B/Base-m, Inc-n’ rule proposed by \cite{de2021continual}, we slice each of the above three datasets into CIFAR-100 B10 Inc10, CIFAR-100 B50 Inc10, CIFAR-100 B5 Inc5, ImageNet-100 B50 Inc10, and Tiny-ImageNet B40 Inc40 for the experiment, where $m$ is the number of classes in the first learning stage, and $n$ represents that of every remaining stage. For a fair comparison, we ensure that the training and test datasets are identical for all methods. Detailed information about the dataset is provided in the Appendix.

\noindent\textbf{Implementation Details.} All methods are implemented in PyTorch, with the baseline methods referencing the PyCIL \cite{zhou2023pycil}, and Gao et al. \cite{gao2025maintaining}. We employ the lightweight ResNet-32 tailored for CIFAR datasets on CIFAR-100, while utilizing ResNet-18 for all other benchmark datasets. The optimizer employs SGD incorporating the multi-step schedule, with a learning rate of 0.1. The detailed training hyperparameters, like the optimizer and batch size, largely follow the default settings in PyCIL. All the methods are run using the same random seed. To eliminate potential confounding effects from increased training epochs, all baselines are run under both the PyCIL-recommended epoch setting and the total epoch budget of DLC's two-stage training, and the maximum accuracy is reported.

\noindent\textbf{Evaluation Metrics.} Following the benchmark protocol settings~\cite{rebuffi2017icarl}, we use $A_b$ to denote the $b$-stage accuracy on the test set that includes all known classes after trained with $D_1, D_2, \cdots, D_b$, $\overline{A} = \frac{1}{T}\sum_{b=1}^{T}A_b$ is average accuracy over $T$ stages, and $A_T$ is the last accuracy on the test set that includes all learned stages. We use $\overline{A}$ and $A_T$ to evaluate the model's performance, which reflects the model's dynamic learning capacity and generalization ability, respectively.
\subsection{Effectiveness on Distillation-Based CL}
\label{sec:52}
\begin{figure}[!t]
  \centering
  \subfloat[iCaRL]{
      \includegraphics[width=1.61in,keepaspectratio]{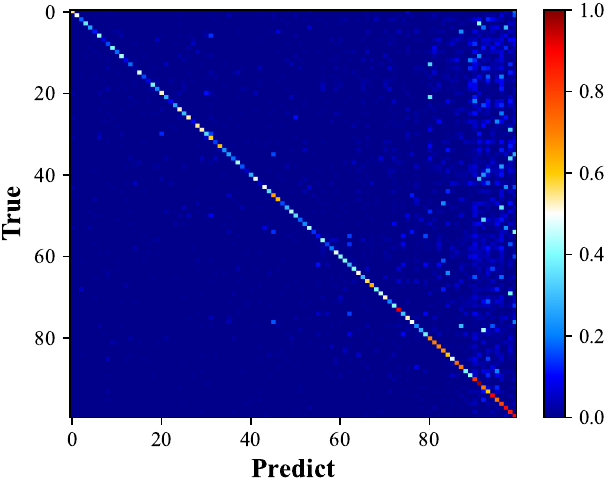}}
   \subfloat[iCaRL-DLC]{
      \includegraphics[width=1.61in,keepaspectratio]{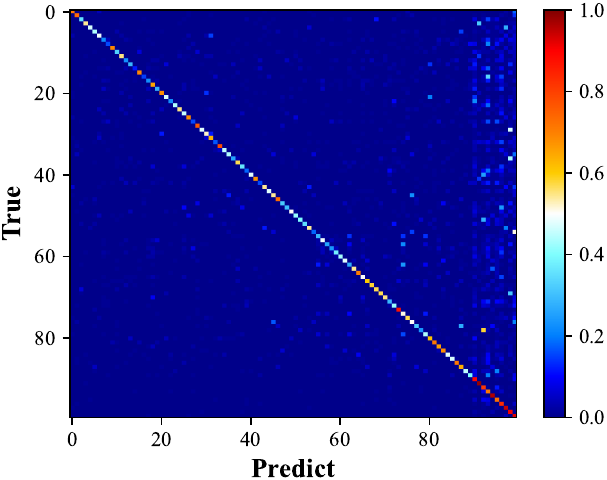}}   
  \caption{Confusion matrix heatmaps of iCaRL with and w/o DLC on CIFAR-100 B10 Inc10.}
  \label{fig:cmap1}
\end{figure}

In this section, we evaluate the performance improvement brought by the proposed DLC framework across different distillation-based CL objectives, including iCaRL \cite{rebuffi2017icarl}, WA \cite{zhao2020maintaining}, BiC \cite{wu2019large}, DAKD \cite{he2024gradient}, and MKD \cite{michel2024rethinking}.  Additionally, a simple herding-based Replay method \cite{rebuffi2017icarl} is included to examine DLC’s contribution in scenarios relying solely on the rehearsal without distillation. We report the $\overline{A}$ and $A_T$ of these established methods with and without DLC across three datasets under varying configurations. In all experiments, the replay buffer size is fixed at 2,000 exemplars. DLC expands only one LoRA plugin per stage, i.e., each plugin consists of a single pair of low-rank matrices $\mathbf{A}$ and $\mathbf{B}$. 
We also report the total parameters required by DLC and the size of the entire feature extractor for comparison. The results are summarized in Table \ref{tab:bc}. The results show that DLC significantly improves all distillation-based methods. Moreover, the parameter overhead introduced by DLC is negligible. DLC achieves up to 8\% accuracy improvement while expanding only 4.8\% to 32\% parameters equivalent to the feature extractor, highlighting its practical value in CL. Fig. \ref{fig:cmap1} shows the confusion matrix heatmap of iCaRL and iCaRL-DLC as an example. DLC increases the thermal intensity along the diagonal for most classes, demonstrating that it could enhance the model's accuracy across most classes. 

\begin{figure}[!t]
  \centering
  \subfloat[B10 Inc10]{
      \includegraphics[width=1.6in,keepaspectratio]{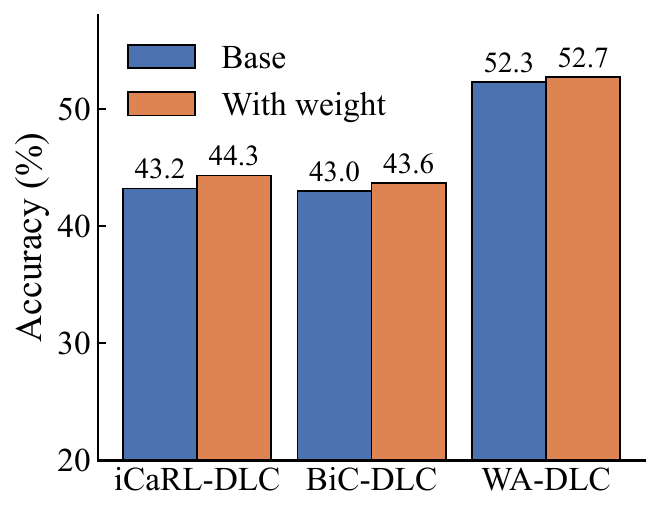}}
   \subfloat[B50 Inc10]{
      \includegraphics[width=1.6in,keepaspectratio]{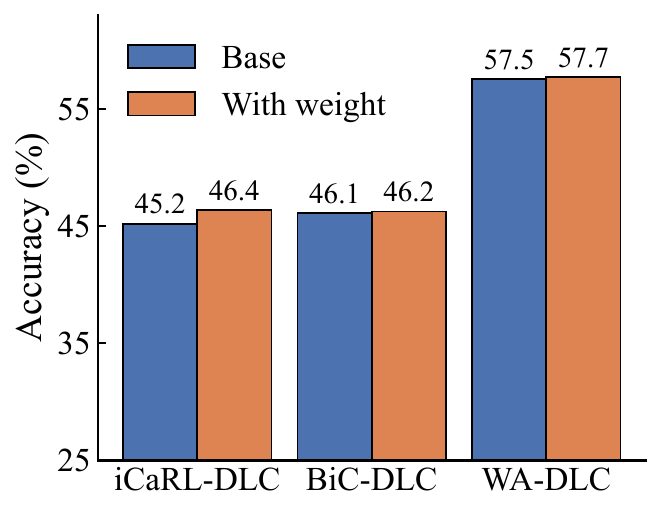}}   
  \caption{Comparison of $A_T$ with and without the weighting unit for DLC-enhanced methods on CIFAR-100.}
  \label{fig:abs1}
\end{figure}

\begin{figure}[!t]
  \centering
  \subfloat[B10 Inc10]{
      \includegraphics[width=1.62in,keepaspectratio]{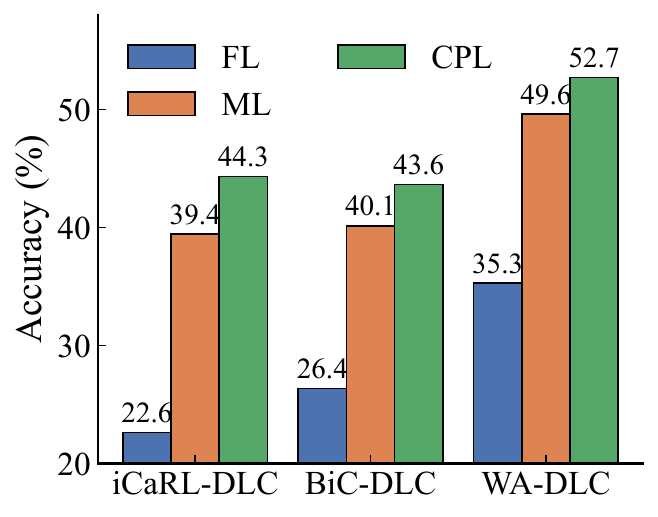}}
   \subfloat[B50 Inc10]{
      \includegraphics[width=1.62in,keepaspectratio]{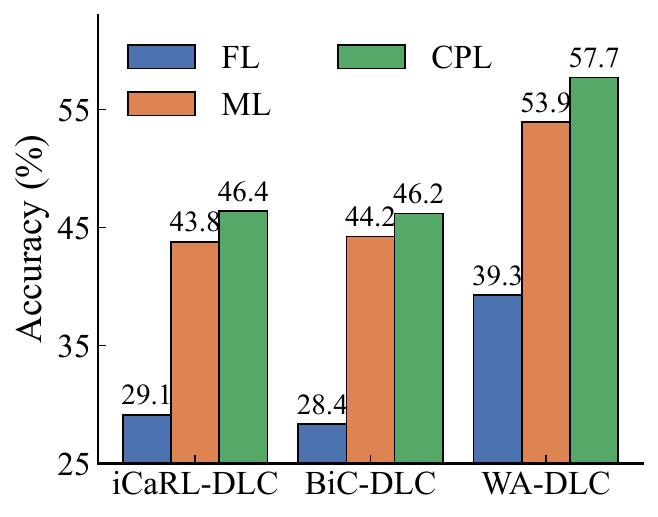}}   
  \caption{Comparison of $A_T$ with plugins deployed at different positions (FL/ML/CPL) on CIFAR-100.}
  \label{fig:abs2}
\end{figure}

In addition to DLC, other plug-and-play enhancement strategies have been proposed to enhance CL methods. For example, some works employ the data augmentation strategy AutoAugment for continual learning, particularly on the CIFAR-100 dataset~\cite{wang2022foster,zheng2025task}. Several plug-and-play loss functions have also been introduced, including $\mathcal{L}_{\text{inter}}$ and $\mathcal{L}_{\text{intra}}$ for distillation methods~\cite{gao2025maintaining}, as well as C-flat~\cite{li2025c}.
 We evaluate the compatibility of DLC with such enhanced strategies. Specifically, we compare the $A_T$ of baseline methods using AutoAugment, the loss  $\mathcal{L}_{\text{inter-intra}}$, and C-flat with and without DLC. For experiments with $\mathcal{L}_{\text{inter-intra}}$, we follow the original implementation and use ResNet-18 as the backbone. As shown in the Fig. \ref{fig:cmp1}, DLC consistently brings significant improvements, demonstrating its compatibility with other plug-and-play enhancements.

Moreover, we conduct two ablation studies. First, we evaluate the effect of the proposed weighting unit by comparing the final accuracy of DLC-enhanced iCaRL, WA, and BiC with and without it, as shown in Fig.~\ref{fig:abs1}. Results indicate that the weighting unit contributes to improved performance. 
Second, we investigate the importance of classifier-proximal deployment by placing the plugin at different depths of the feature extractor, including the first layer (FL), middle layer (ML), and the classifier-proximal layer (CPL).
As shown in Fig.~\ref{fig:abs2}, only classifier-proximal deployment improves performance, whereas earlier deployment causes increasingly severe degradation, indicating poorer distillation compatibility.

\begin{figure*}[!ht]
  \centering
  \subfloat[Long-Sequence]{
      \includegraphics[width=2.2in,keepaspectratio]{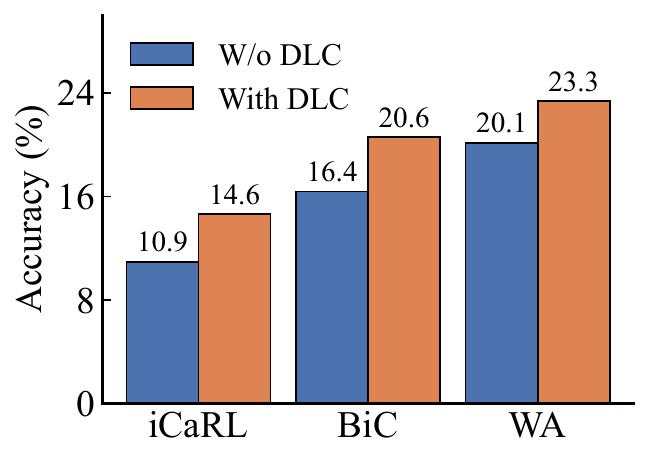}}
  \subfloat[Few-Shot]{
      \includegraphics[width=2.2in,keepaspectratio]{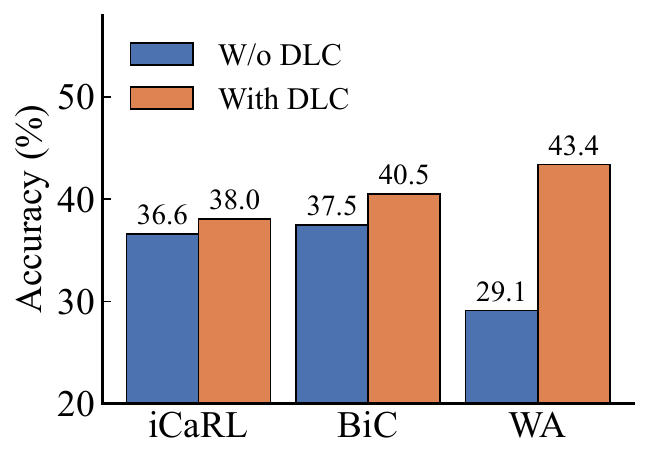}}
  \subfloat[Class-imbalance]{
   \includegraphics[width=2.2in,keepaspectratio]{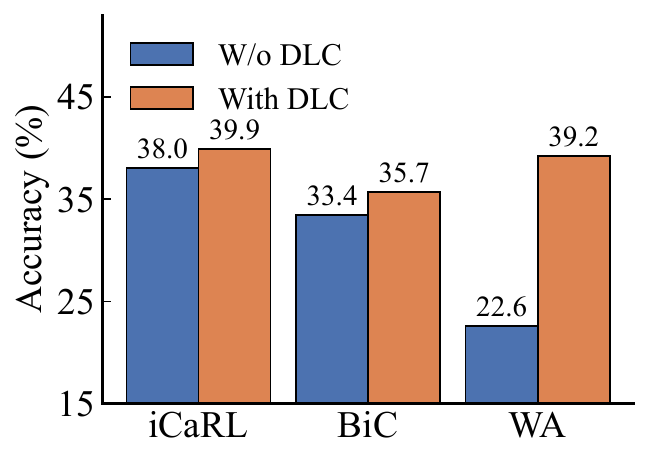}}
  \caption{Comparison of $A_T$ with/without DLC under challenging continual learning settings, including long-sequence, few-shot, and class-imbalanced scenarios.}
  \label{fig:expextreme}
\end{figure*}
\subsection{Evaluation under Challenging CL Settings}
\label{sec:cap}
Distillation-based CL methods are sometimes deployed in extreme scenarios, such as long learning sequences, few-shot, and class-imbalanced data. We therefore further evaluate DLC under these settings. For the long-sequence setting, we use Tiny-ImageNet with the B5Inc5 protocol, resulting in 40 stages. For the few-shot setting, we follow the protocol in~\cite{kim2025new} on CIFAR with B60Inc5, where each class contains only 5 samples after the second stage. For the class-imbalance setting, following Liu et al.~\cite{liu2022long}, we adopt the CIFAR-100 B50Inc5 protocol with an imbalance ratio $\rho = 0.01$ (i.e., head-to-tail ratio takes 100). The results are shown in Fig.~\ref{fig:expextreme}. DLC consistently improves distillation-based CL methods even under these challenging settings.

\begin{figure}[!h]
  \centering
      \includegraphics[width=2.8in,keepaspectratio]{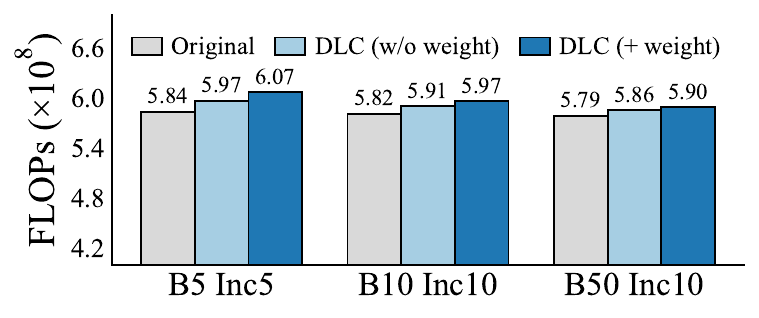}
      
      \includegraphics[width=2.8in,keepaspectratio]{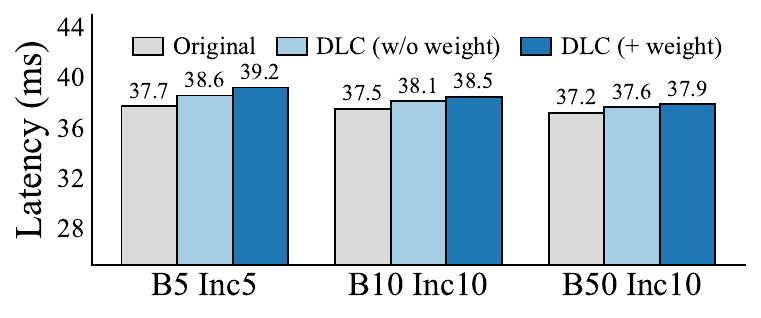}
  \caption{Comparison of inference FLOPs and per-sample inference latency with and w/o DLC.}
  \label{fig:cmptime}
\end{figure}

\subsection{Comparison under Fixed Memory Budget} 
\label{sec:mema}

\begin{table}[!t]
\centering
\caption{Performance comparison on CIFAR-100 with aligned memory cost. `\#P' represents the number of model parameters (million). `\#$\varepsilon$' denotes the number of exemplars, and `MS' denotes the total memory size (MB).}
\scalebox{0.95}{
\begin{tabular}{c||ccccc}
\hline
\multirow{2}{*}{Method} & \multicolumn{5}{c}{CIFAR100 B10 Inc10} \\
\cline{2-6}
 & MS & \#P & \#$\varepsilon$ & $\overline{A}$ & $A_T$ \\
\hline

iCaRL & 23.5 & 0.46 & 7431 & 70.74 & 58.12 \\
WA & 23.5 & 0.46 & 7431 & 69.25 & 59.00 \\
BiC & 23.5 & 0.46 & 7431 & 70.29 & 59.20 \\
FOSTER & 23.5 & 0.46 & 7431 & 72.18 & 59.29 \\
DER+WA & 23.5 & 4.60 & 2000 & 71.24 & 60.26 \\
MEMO+WA & 23.5 & 3.62 & 3312 & 72.32 & 61.92 \\
TagFex+WA & 23.5 & 5.62 & 700 & 74.00 & 64.10 \\
WA-DLC & 23.5 & 0.54 & 7327 & 75.02 & 64.79 \\

\hline
\end{tabular}
}
\label{tab:aligm}
\end{table}

Prior studies have shown that model-level expansion methods can achieve state-of-the-art performance in continual learning, but their gains often rely on substantial and uncounted memory overhead for storing expanded model parameters, making direct comparison with distillation-based methods unfair. To further demonstrate the parameter efficiency and accuracy of DLC under a fair budget constraint, we follow the protocol established in \cite{zhou2022model} and present a memory-aligned comparison of CL baselines.
Under a fixed 23.5 MB memory budget, the memory not consumed by a method's parameters is allocated to replay exemplars. To balance this, methods with fewer parameters store more exemplars, and vice versa. This ensures an equitable evaluation.
In addition to the baselines used in Section \ref{sec:52}, we add several state-of-the-art approaches for comprehensive comparison, including FOSTER \cite{wang2022foster}, DER \cite{yan2021dynamically}, MEMO \cite{zhou2022model}, and TagFex \cite{zheng2025task}. Among them, DER, MEMO, and TagFex have incorporated the weight alignment from WA in their implementations. All methods use the same ResNet backbone. As summarized in Table~\ref{tab:aligm}, WA equipped with DLC achieves the highest classification accuracy under the memory-aligned setting. This highlights that DLC delivers a superior memory-accuracy trade-off.

\subsection{Additional Computational Overhead}
\label{sec:55}
This section investigates the additional inference cost introduced by deploying DLC. Using iCaRL as a representative baseline, we evaluate two deployment variants: iCaRL equipped with only DLC plugins, and iCaRL equipped with both the plugins and the weighting unit. Concretely, on ImageNet-100, we report the total FLOPs and the average inference latency per sample (ms) measured on the same test set. The results are shown in Fig. \ref{fig:cmptime}.
Overall, neither FLOPs nor latency increases substantially as the number of stages grows. Instead, both grow mildly as the stage sequence extends. Even under long sequences such as B5 Inc5, the increases in FLOPs and inference time are only about 4\%.
These results indicate that DLC provides a lightweight and computation-friendly parameter extension for enhancing distillation-based CL.

\section{Conclusion}
In this work, we address a challenge in distillation-based continual learning: improving the model performance with only a small amount of auxiliary parameters, without breaking the single-model distillation setting. We propose DLC, a plug-and-play plugin extension framework that augments replay and distillation-based baselines with LoRA-style residual plugins. DLC deploys each plugin at the classifier-proximal layer and trains it in a decoupled manner, allowing residual correction to directly refine features while remaining compatible with distillation. We further introduce a lightweight weighting unit to down-weight irrelevant features in inference. Extensive experiments demonstrate that DLC consistently improves distillation-based CL baselines with few parameter overhead. Future work will explore more expressive weighting designs for stronger feature integration, as well as earlier-layer plugin deployment to further improve adaptability.


{
    \small
    \bibliographystyle{ieeenat_fullname}
    \bibliography{main}

@inproceedings{yang2024entaugment,
  title={EntAugment: Entropy-Driven Adaptive Data Augmentation Framework for Image Classification},
  author={Yang, Suorong and Shen, Furao and Zhao, Jian},
  booktitle={European Conference on Computer Vision},
  pages={197--214},
  year={2024},
  organization={Springer}
}

@article{gomes2017survey,
  title={A survey on ensemble learning for data stream classification},
  author={Gomes, Heitor Murilo and Barddal, Jean Paul and Enembreck, Fabr{\'\i}cio and Bifet, Albert},
  journal={ACM Computing Surveys (CSUR)},
  volume={50},
  number={2},
  pages={1--36},
  year={2017},
  publisher={ACM New York, NY, USA}
}

@article{de2021continual,
  title={A continual learning survey: Defying forgetting in classification tasks},
  author={De Lange, Matthias and Aljundi, Rahaf and Masana, Marc and Parisot, Sarah and Jia, Xu and Leonardis, Ale{\v{s}} and Slabaugh, Gregory and Tuytelaars, Tinne},
  journal={IEEE transactions on pattern analysis and machine intelligence},
  volume={44},
  number={7},
  pages={3366--3385},
  year={2021},
  publisher={IEEE}
}

@article{rolnick2019experience,
  title={Experience replay for continual learning},
  author={Rolnick, David and Ahuja, Arun and Schwarz, Jonathan and Lillicrap, Timothy and Wayne, Gregory},
  journal={Advances in neural information processing systems},
  volume={32},
  year={2019}
}

@article{nguyen2017variational,
  title={Variational continual learning},
  author={Nguyen, Cuong V and Li, Yingzhen and Bui, Thang D and Turner, Richard E},
  journal={arXiv preprint arXiv:1710.10628},
  year={2017}
}

@inproceedings{mallya2018piggyback,
  title={Piggyback: Adapting a single network to multiple tasks by learning to mask weights},
  author={Mallya, Arun and Davis, Dillon and Lazebnik, Svetlana},
  booktitle={Proceedings of the European conference on computer vision (ECCV)},
  pages={67--82},
  year={2018}
}

@inproceedings{deng2009imagenet,
  title={Imagenet: A large-scale hierarchical image database},
  author={Deng, Jia and Dong, Wei and Socher, Richard and Li, Li-Jia and Li, Kai and Fei-Fei, Li},
  booktitle={2009 IEEE conference on computer vision and pattern recognition},
  pages={248--255},
  year={2009},
  organization={Ieee}
}

@inproceedings{de2021rep,
  title={Continual prototype evolution: Learning online from non-stationary data streams},
  author={De Lange, Matthias and Tuytelaars, Tinne},
  booktitle={Proceedings of the IEEE/CVF international conference on computer vision},
  pages={8250--8259},
  year={2021}
}

@article{lopez2017gradient,
  title={Gradient episodic memory for continual learning},
  author={Lopez-Paz, David and Ranzato, Marc'Aurelio},
  journal={Advances in neural information processing systems},
  volume={30},
  year={2017}
}

@article{ahn2019uncertainty,
  title={Uncertainty-based continual learning with adaptive regularization},
  author={Ahn, Hongjoon and Cha, Sungmin and Lee, Donggyu and Moon, Taesup},
  journal={Advances in neural information processing systems},
  volume={32},
  year={2019}
}

@inproceedings{simon2021learning,
  title={On learning the geodesic path for incremental learning},
  author={Simon, Christian and Koniusz, Piotr and Harandi, Mehrtash},
  booktitle={Proceedings of the IEEE/CVF conference on Computer Vision and Pattern Recognition},
  pages={1591--1600},
  year={2021}
}

@inproceedings{serra2018overcoming,
  title={Overcoming catastrophic forgetting with hard attention to the task},
  author={Serra, Joan and Suris, Didac and Miron, Marius and Karatzoglou, Alexandros},
  booktitle={International conference on machine learning},
  pages={4548--4557},
  year={2018},
  organization={PMLR}
}

@article{xu2018reinforced,
  title={Reinforced continual learning},
  author={Xu, Ju and Zhu, Zhanxing},
  journal={Advances in neural information processing systems},
  volume={31},
  year={2018}
}

@article{zhou2024class,
  title={Class-incremental learning: A survey},
  author={Zhou, Da-Wei and Wang, Qi-Wei and Qi, Zhi-Hong and Ye, Han-Jia and Zhan, De-Chuan and Liu, Ziwei},
  journal={IEEE Transactions on Pattern Analysis and Machine Intelligence},
  year={2024},
  publisher={IEEE}
}

@article{krizhevsky2009learning,
  title={Learning multiple layers of features from tiny images},
  author={Krizhevsky, Alex and Hinton, Geoffrey and others},
  year={2009},
  journal = {Technical report},
  publisher={Toronto, ON, Canada}
}

@inproceedings{rebuffi2017icarl,
  title={icarl: Incremental classifier and representation learning},
  author={Rebuffi, Sylvestre-Alvise and Kolesnikov, Alexander and Sperl, Georg and Lampert, Christoph H},
  booktitle={Proceedings of the IEEE conference on Computer Vision and Pattern Recognition},
  pages={2001--2010},
  year={2017}
}

@inproceedings{wu2019large,
  title={Large scale incremental learning},
  author={Wu, Yue and Chen, Yinpeng and Wang, Lijuan and Ye, Yuancheng and Liu, Zicheng and Guo, Yandong and Fu, Yun},
  booktitle={Proceedings of the IEEE/CVF conference on computer vision and pattern recognition},
  pages={374--382},
  year={2019}
}

@inproceedings{zhao2020maintaining,
  title={Maintaining discrimination and fairness in class incremental learning},
  author={Zhao, Bowen and Xiao, Xi and Gan, Guojun and Zhang, Bin and Xia, Shu-Tao},
  booktitle={Proceedings of the IEEE/CVF conference on computer vision and pattern recognition},
  pages={13208--13217},
  year={2020}
}

@inproceedings{yan2021dynamically,
  title={Der: Dynamically expandable representation for class incremental learning},
  author={Yan, Shipeng and Xie, Jiangwei and He, Xuming},
  booktitle={Proceedings of the IEEE/CVF conference on computer vision and pattern recognition},
  pages={3014--3023},
  year={2021}
}

@inproceedings{wang2022foster,
  title={Foster: Feature boosting and compression for class-incremental learning},
  author={Wang, Fu-Yun and Zhou, Da-Wei and Ye, Han-Jia and Zhan, De-Chuan},
  booktitle={European conference on computer vision},
  pages={398--414},
  year={2022},
  organization={Springer}
}

@article{zhou2022model,
  title={A model or 603 exemplars: Towards memory-efficient class-incremental learning},
  author={Zhou, Da-Wei and Wang, Qi-Wei and Ye, Han-Jia and Zhan, De-Chuan},
  journal={arXiv preprint arXiv:2205.13218},
  year={2022}
}

@article{liangloss,
  title={Loss decoupling for task-agnostic continual learning},
  author={Liang, Yan-Shuo and Li, Wu-Jun},
  journal={Advances in Neural Information Processing Systems},
  volume={36},
  pages={11151--11167},
  year={2024}
}

@article{bian2024make,
  title={Make continual learning stronger via c-flat},
  author={Bian, Ang and Li, Wei and Yuan, Hangjie and Wang, Mang and Zhao, Zixiang and Lu, Aojun and Ji, Pengliang and Feng, Tao and others},
  journal={Advances in Neural Information Processing Systems},
  volume={37},
  pages={7608--7630},
  year={2024}
}

@inproceedings{zhou2024expandable,
  title={Expandable subspace ensemble for pre-trained model-based class-incremental learning},
  author={Zhou, Da-Wei and Sun, Hai-Long and Ye, Han-Jia and Zhan, De-Chuan},
  booktitle={Proceedings of the IEEE/CVF Conference on Computer Vision and Pattern Recognition},
  pages={23554--23564},
  year={2024}
}

@article{yang2025supervised,
  title={Supervised contrastive learning with prototype distillation for data incremental learning},
  author={Yang, Suorong and Zhang, Tianyue and Xu, Zhiming and Li, Peijia and Xu, Baile and Shen, Furao and Zhao, Jian},
  journal={Neural Networks},
  pages={107651},
  year={2025},
  publisher={Elsevier}
}

@inproceedings{gao2025maintaining,
  title={Maintaining fairness in logit-based knowledge distillation for class-incremental learning},
  author={Gao, Zijian and Han, Shanhao and Zhang, Xingxing and Xu, Kele and Zhou, Dulan and Mao, Xinjun and Dou, Yong and Wang, Huaimin},
  booktitle={Proceedings of the AAAI Conference on Artificial Intelligence},
  volume={39},
  pages={16763--16771},
  year={2025}
}

@inproceedings{sun2025mos,
  title={Mos: Model surgery for pre-trained model-based class-incremental learning},
  author={Sun, Hai-Long and Zhou, Da-Wei and Zhao, Hanbin and Gan, Le and Zhan, De-Chuan and Ye, Han-Jia},
  booktitle={Proceedings of the AAAI Conference on Artificial Intelligence},
  volume={39},
  pages={20699--20707},
  year={2025}
}

@article{vinyals2016matching,
  title={Matching networks for one shot learning},
  author={Vinyals, Oriol and Blundell, Charles and Lillicrap, Timothy and Wierstra, Daan and others},
  journal={Advances in neural information processing systems},
  volume={29},
  year={2016}
}

@inproceedings{aghasanli2025prototype,
  title={Prototype-Based Continual Learning with Label-free Replay Buffer and Cluster Preservation Loss},
  author={Aghasanli, Agil and Li, Yi and Angelov, Plamen},
  booktitle={Proceedings of the Computer Vision and Pattern Recognition Conference},
  pages={6545--6554},
  year={2025}
}

@inproceedings{zheng2025task,
  title={Task-Agnostic Guided Feature Expansion for Class-Incremental Learning},
  author={Zheng, Bowen and Zhou, Da-Wei and Ye, Han-Jia and Zhan, De-Chuan},
  booktitle={Proceedings of the Computer Vision and Pattern Recognition Conference},
  pages={10099--10109},
  year={2025}
}

@article{hu2022lora,
  title={Lora: Low-rank adaptation of large language models.},
  author={Hu, Edward J and Shen, Yelong and Wallis, Phillip and Allen-Zhu, Zeyuan and Li, Yuanzhi and Wang, Shean and Wang, Lu and Chen, Weizhu and others},
  journal={ICLR},
  volume={1},
  number={2},
  pages={3},
  year={2022}
}

@misc{zhou2023pycil,
  title={Pycil: a python toolbox for class-incremental learning},
  author={Zhou, Da-Wei and Wang, Fu-Yun and Ye, Han-Jia and Zhan, De-Chuan},
  year={2023},
  publisher={Springer}
}

@article{li2025re,
  title={Re-fed+: A better replay strategy for federated incremental learning},
  author={Li, Yichen and Wang, Haozhao and Qi, Yining and Liu, Wei and Li, Ruixuan},
  journal={IEEE Transactions on Pattern Analysis and Machine Intelligence},
  year={2025},
  publisher={IEEE}
}

@inproceedings{he2024gradient,
  title={Gradient reweighting: Towards imbalanced class-incremental learning},
  author={He, Jiangpeng},
  booktitle={Proceedings of the IEEE/CVF Conference on Computer Vision and Pattern Recognition},
  pages={16668--16677},
  year={2024}
}

@article{zeng2025terrasap,
  title={TerraSAP: Spatially Aware Prompt-Based Framework for Few-Shot Class-Incremental Learning in Remote Sensing Image Classification},
  author={Zeng, Jiaxing and Tan, Yifeng and Yang, Lina and Zhang, Siwei and Liang, Lianhui},
  journal={IEEE Journal of Selected Topics in Applied Earth Observations and Remote Sensing},
  volume={19},
  pages={3143--3156},
  year={2025},
  publisher={IEEE}
}

@article{nori2025federated,
  title={Federated class-incremental learning: A hybrid approach using latent exemplars and data-free techniques to address local and global forgetting},
  author={Nori, Milad Khademi and Kim, Il-Min and Wang, Guanghui},
  journal={arXiv preprint arXiv:2501.15356},
  year={2025}
}

@inproceedings{michel2024rethinking,
  title={Rethinking Momentum Knowledge Distillation in Online Continual Learning},
  author={Michel, Nicolas and Wang, Maorong and Xiao, Ling and Yamasaki, Toshihiko},
  booktitle={International Conference on Machine Learning},
  pages={35607--35622},
  year={2024},
  organization={PMLR}
}

@inproceedings{wang2024improving,
  title={Improving plasticity in online continual learning via collaborative learning},
  author={Wang, Maorong and Michel, Nicolas and Xiao, Ling and Yamasaki, Toshihiko},
  booktitle={Proceedings of the IEEE/CVF Conference on Computer Vision and Pattern Recognition},
  pages={23460--23469},
  year={2024}
}

@inproceedings{de2024towards,
  title={Towards trustworthy MetaShopping: Studying manipulative audiovisual designs in virtual-physical commercial platforms},
  author={De Haas, Esm{\'e}e Henrieke Anne and Lee, Lik-Hang and Huang, Yiming and Bermejo, Carlos and Hui, Pan and Lin, Zijun},
  booktitle={Proceedings of the 32nd ACM International Conference on Multimedia},
  pages={68--77},
  year={2024}
}

@inproceedings{gui2024navigating,
  title={Navigating weight prediction with diet diary},
  author={Gui, Yinxuan and Zhu, Bin and Chen, Jingjing and Ngo, Chong Wah and Jiang, Yu-Gang},
  booktitle={Proceedings of the 32nd ACM International Conference on Multimedia},
  pages={127--136},
  year={2024}
}

@inproceedings{wu2025gui,
  title={GUI-Narrator: Detecting and Captioning Computer GUI Actions},
  author={Wu, Qinchen and Gao, Difei and Lin, Qinghong and Wu, Zhuoyu and Shou, Mike Zheng},
  booktitle={Proceedings of the 33rd ACM International Conference on Multimedia},
  pages={3683--3692},
  year={2025}
}

@inproceedings{ye2025safedriverag,
  title={Safedriverag: Towards safe autonomous driving with knowledge graph-based retrieval-augmented generation},
  author={Ye, Hao and Qi, Mengshi and Liu, Zhaohong and Liu, Liang and Ma, Huadong},
  booktitle={Proceedings of the 33rd ACM International Conference on Multimedia},
  pages={11170--11178},
  year={2025}
}

@article{li2025c,
  title={C-flat++: Towards a more efficient and powerful framework for continual learning},
  author={Li, Wei and Yuan, Hangjie and Zhao, Zixiang and Zhu, Yifan and Lu, Aojun and Feng, Tao and Sun, Yanan},
  journal={arXiv preprint arXiv:2508.18860},
  year={2025}
}

@article{hayou2024lora+,
  title={Lora+: Efficient low rank adaptation of large models},
  author={Hayou, Soufiane and Ghosh, Nikhil and Yu, Bin},
  journal={arXiv preprint arXiv:2402.12354},
  year={2024}
}

@article{han2024parameter,
  title={Parameter-efficient fine-tuning for large models: A comprehensive survey},
  author={Han, Zeyu and Gao, Chao and Liu, Jinyang and Zhang, Jeff and Zhang, Sai Qian},
  journal={arXiv preprint arXiv:2403.14608},
  year={2024}
}

@inproceedings{huang2017densely,
  title={Densely connected convolutional networks},
  author={Huang, Gao and Liu, Zhuang and Van Der Maaten, Laurens and Weinberger, Kilian Q},
  booktitle={Proceedings of the IEEE conference on computer vision and pattern recognition},
  pages={4700--4708},
  year={2017}
}

@article{kim2025new,
  title={A New Benchmark for Few-Shot Class-Incremental Learning: Redefining the Upper Bound},
  author={Kim, Shiwon and Hwang, Dongjun and Woo, Sungwon and Singh, Rita},
  journal={arXiv e-prints},
  pages={arXiv--2503},
  year={2025}
}

@inproceedings{liu2022long,
  title={Long-tailed class incremental learning},
  author={Liu, Xialei and Hu, Yu-Song and Cao, Xu-Sheng and Bagdanov, Andrew D and Li, Ke and Cheng, Ming-Ming},
  booktitle={European Conference on Computer Vision},
  pages={495--512},
  year={2022},
  organization={Springer}
}
}


\end{document}